\documentclass[journal,twoside,web]{ieeecolor}
\usepackage{generic}
\usepackage[table]{xcolor}
\usepackage{cite}
\usepackage{amsmath,amssymb,amsfonts}
\usepackage{algorithmic}
\usepackage{graphicx}
\usepackage{algorithm,algorithmic}
\usepackage{hyperref}
\hypersetup{hidelinks=true}
\usepackage{textcomp}
\usepackage{arydshln}
\usepackage{booktabs}
\usepackage{multirow}
\usepackage{subfig}
\usepackage{booktabs}
\usepackage{orcidlink}
\def\BibTeX{{\rm B\kern-.05em{\sc i\kern-.025em b}\kern-.08em
    T\kern-.1667em\lower.7ex\hbox{E}\kern-.125emX}}
\begin{document}
\title{EEG-VLM: A Hierarchical Vision-Language Model with Multi-Level Feature Alignment and Visually Enhanced Language-Guided Reasoning for EEG Image-Based Sleep Stage Prediction}

\author{
Xihe Qiu\orcidlink{0000-0003-4024-925X}
Gengchen Ma\orcidlink{0000-0001-7511-2910},
Haoyu Wang\orcidlink{0000-0001-9575-7345}, 
Chen Zhan\orcidlink{0009-0003-1911-0242}, 
Xiaoyu Tan*\orcidlink{0000-0003-3555-7143},
and Shuo Li\orcidlink{0000-0002-5184-3230}
\thanks{This work is supported by the Shanghai Municipal Natural Science Foundation (No.23ZR1425400). \emph{(*Corresponding Author: Xiaoyu Tan(arthurtan@tencent.com).)}.}
\thanks{Xihe Qiu and Gengchen Ma contributed equally to this work.} 
\thanks{Xihe Qiu, Gengchen Ma and Chen Zhan are with School of Electronic and Electrical Engineering, Shanghai University of Engineering Science, Shanghai, 201620, China.}
\thanks{Haoyu Wang is with the Department of Control Science and Engineering, College of Electronics and Information Engineering, Tongji University, Shanghai 200092, China.}
\thanks{Xiaoyu Tan is with Tencent Youtu Lab, Shanghai 200232, China, China.}
\thanks{Shuo Li is with Case Western Reserve University, USA.}
}

\maketitle
\begin{abstract}
Sleep stage classification based on electroencephalography (EEG) is fundamental for assessing sleep quality and diagnosing sleep-related disorders. However, most traditional machine learning methods rely heavily on prior knowledge and handcrafted features, while existing deep learning models still struggle to jointly capture fine-grained time–frequency patterns and achieve clinical interpretability. Recently, vision–language models (VLMs) have made significant progress in the medical domain, yet their performance remains constrained when applied to physiological waveform data, especially EEG signals, due to their limited visual understanding and insufficient reasoning capability. To address these challenges, we propose \textit{EEG-VLM}, a hierarchical vision–language framework that integrates multi-level feature alignment with visually enhanced language-guided reasoning for interpretable EEG-based sleep stage classification. Specifically, a specialized visual enhancement module constructs high-level visual tokens from intermediate-layer features to extract rich semantic representations of EEG images. These tokens are further aligned with low-level CLIP features through a multi-level alignment mechanism, enhancing the VLM’s image-processing capability. In addition, a Chain-of-Thought (CoT) reasoning strategy decomposes complex medical inference into interpretable logical steps, effectively simulating expert-like decision-making. Experimental results demonstrate that the proposed method significantly improves both the accuracy and interpretability of VLMs in EEG-based sleep stage classification, showing promising potential for automated and explainable EEG analysis in clinical settings.
\end{abstract}


\section{Introduction}
\label{sec:introduction}
Sleep plays a vital role in maintaining brain function and overall physiological health \cite{czeisler2015duration}. Accurate assessment of sleep quality not only reflects an individual’s health status but also provides an essential basis for diagnosing and treating sleep-related disorders \cite{vatankhah2010intelligent,brignol2012eeg,zhu2014analysis}. Currently, the American Academy of Sleep Medicine (AASM) standards \cite{berry2012aasm} are widely adopted for sleep stage scoring. Among various physiological signals, electroencephalography (EEG) is widely regarded as the most informative and commonly used modality for sleep stage classification \cite{kayikcioglu2015fast,alickovic2018ensemble,an2021unsupervised}, as it captures rich physiological and pathological information and clearly differentiates between different sleep stages \cite{li2015development,manjunath2024detection}.

Waveform morphology and frequency composition are central to EEG-based sleep stage classification. Sleep experts rely on identifying characteristic waveforms—such as alpha, beta, and theta rhythms—within each 30-second epoch to determine sleep stages. However, sleep stage classification is guided by complex clinical criteria, making it a labor-intensive, time-consuming process that is prone to inter-rater variability.

To overcome these limitations, numerous automatic sleep stage classification methods have been proposed. Traditional machine learning approaches \cite{phan2013metric,seifpour2018new,satapathy2022effective,arslan2023sleep} rely heavily on handcrafted features and prior domain knowledge, which limits their adaptability and generalization capability. In contrast, deep learning-based methods \cite{nie2021recsleepnet,eldele2021attention,zhang2023shnn,pham2023automatic} can automatically extract discriminative representations from EEG signals. However, they often struggle to capture fine-grained distinctions—particularly between physiologically similar stages such as N1 and REM—resulting in suboptimal performance.

Recently, vision–language models (VLMs) \cite{GPT-4,liu2023llava,liu2024llavanext,Qwen-VL,Qwen2VL,llama3.2vision} have demonstrated impressive performance across various general-purpose tasks by leveraging joint visual–textual representations. Although their application in the medical domain has gained increasing attention, their performance remains limited when applied to physiological waveform data—particularly EEG—due to insufficient capacity for fine-grained visual perception, effective image processing, and domain-specific reasoning \cite{wu2023can,abdullahi2024learning,kaczmarczyk2024evaluating}. These challenges restrict the applicability of VLMs in complex clinical contexts such as EEG-based sleep stage classification.

To address these challenges, we propose a hierarchical vision–language framework tailored for EEG image-based representation learning. Specifically, we augment the visual encoder with a visual enhancement module that extracts intermediate-level representations and transforms them into high-level visual tokens, enabling the model to capture both fine-grained visual details and abstract semantic information. These high-level semantic representations are then aligned and integrated with low-level visual features extracted by CLIP through a multi-level feature alignment mechanism, facilitating multi-scale perception and bridging semantic gaps across hierarchical representations. On the language side, we incorporate a CoT prompting strategy to guide the model through structured, step-wise reasoning that simulates expert decision-making. This integrated architecture allows the model to achieve accurate and interpretable predictions, particularly for ambiguous stages such as N1 and REM.

The key contributions of our work are summarized as follows.
\begin{enumerate}
    \item We propose EEG-VLM, a hierarchical VLM that combines multi-level feature alignment with visually enhanced language-guided reasoning, demonstrating the feasibility and potential of VLM for EEG-based sleep stage classification.
    \item We design a visual enhancement module that constructs high-level visual representations from intermediate-layer features, enabling the model to capture deep semantic information from EEG images.
    \item We introduce a multi-level feature alignment mechanism to effectively fuse visual tokens from different levels, thereby enhancing the model's image processing and feature representation capabilities.
    \item By employing CoT reasoning, we simplify complex inference tasks, improving the transparency and accuracy of the model's decision-making while effectively simulating the step-by-step judgment of human experts.
    \item Experimental results show that our method demonstrates robustness to ambiguous sleep stage boundaries (\textit{e.g.}, N1 and REM) and enhanced interpretability, providing new insights and a promising direction for physiological waveform analysis.
\end{enumerate}

\section{Related Work}
\subsection{Traditional and Deep Learning for EEG Sleep Stage Classification}

Traditional approaches to automatic sleep stage classification primarily rely on handcrafted features extracted from the time, frequency, or time-frequency domains of EEG signals. These features are typically fed into classical machine learning algorithms such as Support Vector Machines (SVM), k-Nearest Neighbors (KNN), or Random Forests (RF) \cite{alickovic2018ensemble,aboalayon2016sleep}. For example, \cite{hassan2017automated} combined tunable-Q factor wavelet transform (TQWT) and normal inverse Gaussian (NIG) parameters with AdaBoost, while \cite{jiang2019robust} integrated multiple signal decomposition and feature extraction methods with rule-free hidden Markov model (HMM) refinement for single-channel EEG. Although these methods can achieve reasonable accuracy, they often suffer from limited generalizability and require extensive domain expertise for feature engineering.

In recent years, deep learning has enabled end-to-end models capable of learning hierarchical representations directly from raw EEG signals. Architectures based on Convolutional Neural Networks (CNNs), Recurrent Neural Networks (RNNs), and Transformers have achieved state-of-the-art performance. For instance, DeepSleepNet~\cite{supratak2017deepsleepnet} adopted a hybrid CNN-RNN architecture to automatically extract time-invariant features and model temporal dependencies from single-channel EEG, while~\cite{phan2018joint} proposed a joint classification-prediction CNN to exploit sequential context. More recent designs, such as SleepEEGNet~\cite{mousavi2019sleepeegnet}, incorporate attention mechanisms and multi-resolution processing to better capture temporal and spectral dynamics. Nevertheless, these models still face challenges in distinguishing physiologically similar stages like N1 and REM, due to subtle and overlapping signal characteristics. Furthermore, their limited interpretability and underutilization of frequency-domain priors constrain clinical trust and deployment, especially in borderline or pathological cases.

\subsection{Vision-Language Models in the Medical Domain: Opportunities and Challenges}
Recent advances in large-scale multimodal models, such as ChatGPT, LLaVA, and Qwen-VL, have significantly advanced the field of VLMs~\cite{liu2023llava, GPT-4, Qwen2VL}. These models have achieved state-of-the-art performance in tasks including image captioning, visual question answering (VQA), and multimodal reasoning. Increasingly, VLMs are being adapted for medical applications such as radiology report generation, digital pathology, and biomedical image analysis~\cite{radford2021learning,li2022blip,liang2024survey,lu2024visual}. 

However, the application of VLMs to physiological waveform data—particularly EEG—remains underexplored. The high visual complexity of EEG images limit the effectiveness of generic VLMs (\textit{e.g.}, CLIP)~\cite{ferrante2024decoding}. Current models struggle to capture fine-grained details essential for clinical interpretation and lack the domain-specific inductive biases and interpretability required in high-stakes medical scenarios. This limitation is particularly evident in EEG-based sleep stage classification, where robust visual perception and clinically interpretable reasoning are critical for real-world deployment~\cite{stiglic2020interpretability}.

\begin{figure*}[ht]
\centering
\includegraphics[width=\textwidth]{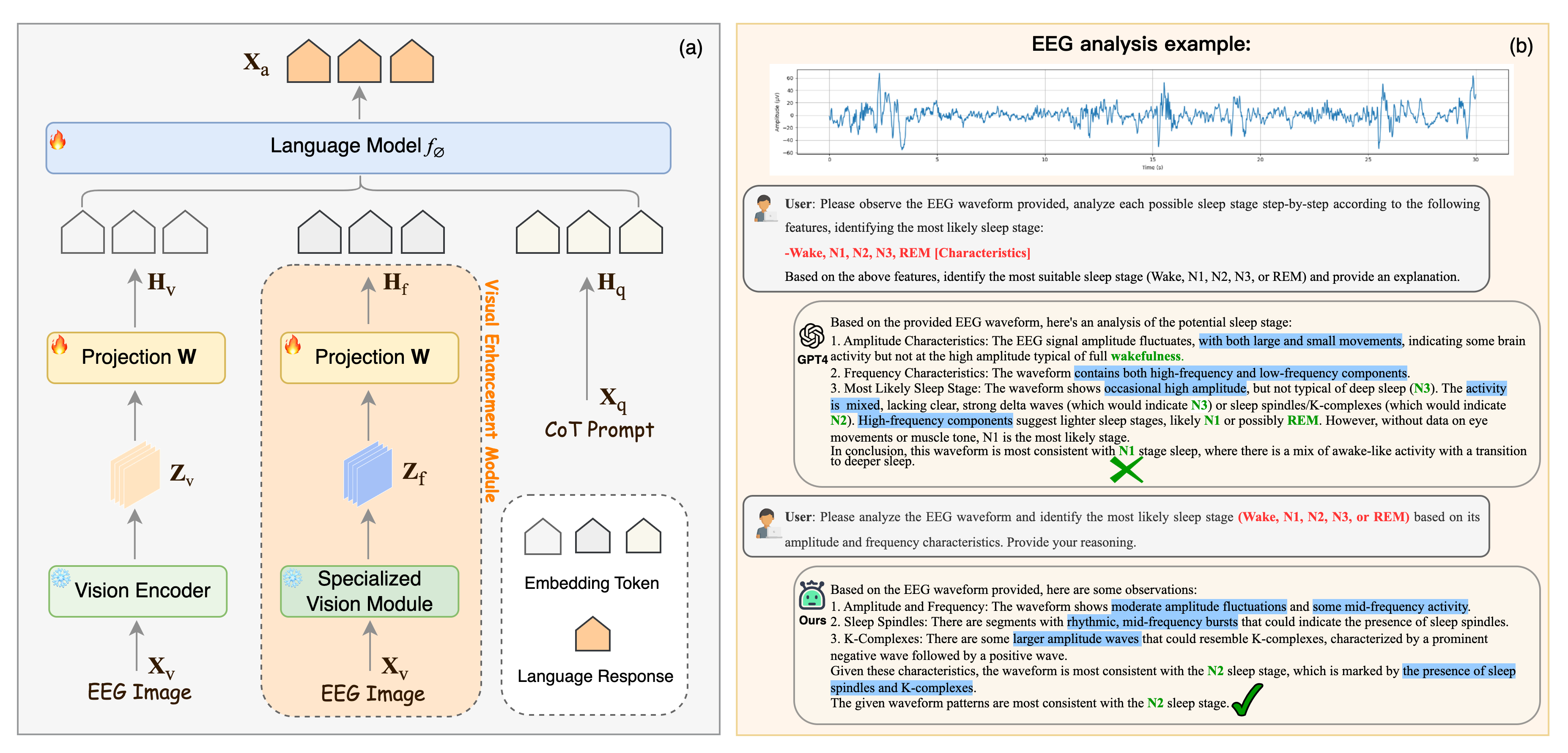}
\caption{Model Architecture and EEG Analysis Example: (a) Proposed Model Architecture; (b) EEG Analysis Example.} \label{main}
\end{figure*}

\section{Methodology}
\subsection{Overview}
The proposed method integrates a vision encoder, a language model, and a visual enhancement module, as illustrated in Fig.~\ref{main}(a). To balance performance and computational efficiency, we adopt the LLaVA-1.5-13B model as the backbone to evaluate the effectiveness of our framework.

In our framework, the input consists of an EEG image \( X_v \) and a CoT prompt \cite{wei2022chain} \( X_q \). The EEG image \( X_v \) is first processed by a pre-trained CLIP vision encoder (ViT-L/14) \cite{radford2021learning} to extract low-level visual features \( Z_v = g(X_v) \). Simultaneously, the image is passed through a specialized vision module to obtain high-level semantic features \( Z_f = \psi(X_v) \). These hierarchical representations, \( Z_v \) and \( Z_f \), are transformed into language embedding tokens \( H_v \) and \( H_f \) through a shared projection layer \( W \), consisting of two MLP layers, as follows:
\begin{equation}
H_v = W \cdot Z_v, \quad H_f = W \cdot Z_f, \label{eq:eq1}
\end{equation}
where $\quad Z_v = g(X_v), \quad Z_f = \psi(X_v)$. Then, \( H_f \) is passed through the multi-level feature alignment function \( H(\cdot) \) to generate the final feature embedding token \( H_f' = H(H_f) \), which is then passed, along with the visual embedding token \( H_v \) and the text embedding token \( H_q \) derived from processing the CoT prompt, into the language model \( f_\phi \) to generate the final language response \( X_a \):
\begin{equation}
H_f' = H(H_f), X_a = f(H_v, H_f', H_q)\label{eq:eq2}
\end{equation}
\subsection{Visual Enhancement Module}
Prior studies such as LLaVolta \cite{chen2024efficient} indicate that VLMs still struggle to effectively represent and process visual information. The intricate details of EEG pose significant challenges for VLMs when handling such tasks. To overcome this limitation, we designed a visual enhancement module, in which a specialized vision module is responsible for capturing high-level semantic representations from EEG images, thereby enhancing the VLM's visual understanding and processing capabilities.

In our benchmark setting, the specialized vision module is implemented using a modified ResNet-18 \cite{he2016deep} architecture. The modifications are as follows:

\textbf{Modification to the final convolutional layer}: The output channels of the last convolutional layer are increased from 512 to 1024 to align with the dimensionality of the low-level visual features \( Z_v \). 

\textbf{Adjustment to the batch normalization layer}: The batch normalization layer is updated to match the new output channel size of 1024.

\textbf{Addition of a 1x1 convolution in the downsampling component}: To ensure channel size matching between residual connections, a 1x1 convolution is added to the downsampling component, increasing the input channels from 512 to 1024.  

\textbf{Fully connected layer for classification}: After modification, the feature map is flattened and passed through a fully connected layer for classification.  
\begin{equation}
y = W \cdot \text{Flattened Features}\label{classification}
\end{equation}

These modifications produce intermediate features \( Z_f \) that are spatially aligned with the low-level visual features \( Z_v \) and retain both fine-grained details and global semantics, making them suitable for alignment with text or other modalities. These intermediate features (\( Z_f \)), extracted immediately before the classification layer, are converted into fixed-dimension tokens and fed into the VLM for subsequent processing.

Finally, \( Z_f \) is passed through a shared mapping layer to generate the preliminary feature embedding token \( H_f \), which is subsequently used to enhance the visual representation and understanding capabilities of the VLMs.

\begin{figure*}[h]
\includegraphics[width=\textwidth]{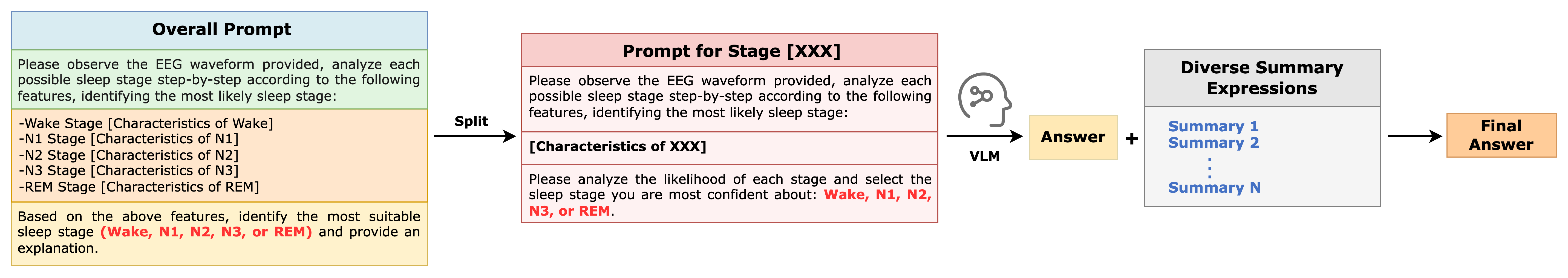}
\caption{CoT-Guided Multi-Step EEG Sleep Stage Analysis Generation} \label{cot}
\end{figure*}

\subsection{Multi-Level Feature Alignment}
Through the aforementioned method, we obtain hierarchical feature embeddings \( H_v \) and \( H_f \). However, how to effectively construct hierarchical embedding tokens to enhance the visual processing capabilities of the VLMs remains a challenge. To address this, we propose the following approach:
\begin{equation}
H'_f = H_v + \text{Expand}(H_f)\label{H}
\end{equation}
where \( \text{Expand}(H_f) \) replicates \( H_f \) along the patch dimension to match the size of \( H_v \). This expanded \( H_f \) is then added element-wise to \( H_v \) to produce the final feature embedding token \( H'_f \). This operation defines the multi-level feature alignment function \( H(\cdot) \).

This method enables the model to process local regions while integrating fine-grained visual information and global semantic priors, thereby enhancing its ability to represent features and process EEG images.

\subsection{Stage-Wise CoT Generation for EEG Sleep Stage Classification}
Although hierarchical representation learning significantly enhances the visual understanding capabilities of VLMs, their performance on complex clinical reasoning tasks remains limited. This limitation is especially evident in EEG sleep stage classification, where subtle physiological differences—particularly between stages such as N1 and REM—require expert-like, stage-specific judgment.

To address this issue, we propose a Stage-Wise CoT prompting strategy to guide a powerful VLM (\textit{e.g.}, GPT-4) in generating high-quality CoT data, simulating the thought process of human experts.

Specifically, our strategy breaks down the global sleep stage analysis task into a series of focused, interpretable sub-tasks, as illustrated in Fig.~\ref{cot}. Rather than directly prompting a VLM with an overall CoT instruction—which often results in vague or inconsistent outputs, as shown in Fig.~\ref{main}(b)—we decompose the task into sub-CoT prompts, each tailored to a specific sleep stage (\textit{e.g.}, Wake, N1, N2, N3, REM). Each prompt emphasizes the relevant waveform features and frequency–amplitude patterns, enabling the model to conduct targeted, stage-specific reasoning.

Each of these sub-prompts is processed independently by the VLM to generate preliminary stage-level analyses. To further enhance the consistency and robustness of the output, we combine the model’s intermediate answers with diverse summary expressions to construct a coherent and interpretable final answer.

This Stage-Wise CoT strategy not only improves the quality and interpretability of the generated data—especially for ambiguous stages—but also more closely simulates the step-by-step analytical process of human experts, thereby enhancing both the transparency and clinical reliability of the model’s decision-making process.

\begin{table*}[t]
\centering
\renewcommand{\arraystretch}{1.15}
\setlength{\tabcolsep}{7pt}

\caption{Performance Comparison of Different Approaches on the Sleep-EDFx Dataset. 
\textbf{Bold} indicates the best result and \underline{underline} indicates the second-best result.}
\label{tab:Main}

\rowcolors{3}{}{gray!8}

\begin{tabular}{c ccc ccccc}
\toprule

\textbf{Method} &
\multicolumn{3}{c}{\textbf{Overall Results}} &
\multicolumn{5}{c}{\textbf{F1-score for Each Class}} \\
\cmidrule(lr){2-4} \cmidrule(lr){5-9}

 & \textbf{Accuracy} & \textbf{MF1} & \textbf{Kappa} &
   \textbf{Wake} & \textbf{N1} & \textbf{N2} & \textbf{N3} & \textbf{REM} \\
\midrule

GPT-4             & 0.205 & 0.197 & 0.007 & 0.170 & 0.167 & 0.217 & 0.168 & 0.265 \\
Qwen2.5-VL-72B     & 0.243 & 0.179 & 0.053 & 0.035 & 0.324 & 0.367 & 0.000 & 0.169 \\
LLaVA-Next-8B      & 0.533 & 0.483 & 0.417 & 0.655 & 0.075 & 0.417 & 0.760 & 0.510 \\
FFTCN              & \textbf{0.826} & 0.771 & \underline{0.760} & \textbf{0.922} & 0.473 & 0.848 & 0.800 & \textbf{0.810} \\
ResNet-18          & 0.752 & 0.756 & 0.690 & 0.795 & 0.637 & 0.842 & \underline{0.937} & 0.567 \\
ConvNeXt-Base      & \underline{0.813} & \textbf{0.818} & \underline{0.760} & 0.835 & \underline{0.715} & \textbf{0.876} & 0.905 & 0.761 \\

\rowcolor{white}
\textbf{Ours-L1.5-R18}   & 0.792 & 0.797 & 0.740 & 0.839 & 0.654 & \underline{0.859} & \textbf{0.944} & 0.688 \\
\textbf{Ours-LNxt-R18}   & 0.787 & 0.793 & 0.734 & 0.814 & 0.682 & 0.838 & 0.931 & 0.702 \\
\textbf{Ours-L1.5-CNx}   & 0.811 & \underline{0.816} & \textbf{0.763} & \underline{0.851} & \textbf{0.717} & 0.846 & 0.905 & 0.760 \\
\textbf{Ours-LNxt-CNx}   & 0.808 & 0.813 & \underline{0.760} & 0.849 & 0.694 & 0.855 & 0.897 & \underline{0.771} \\

\bottomrule
\end{tabular}
\end{table*}

\begin{figure*}[tb]
\centering
\includegraphics[width=0.90\textwidth]{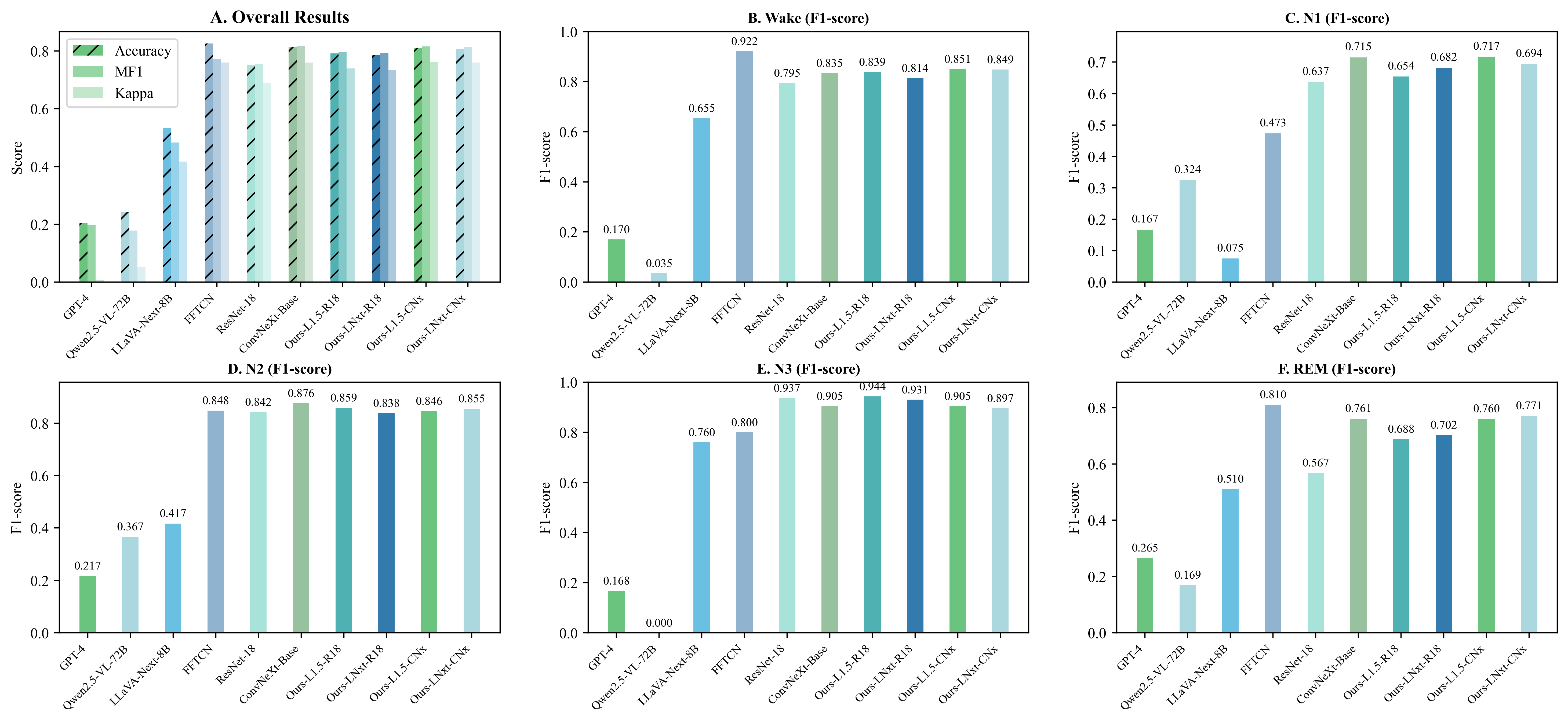}
\caption{Overall and Per-Stage Classification Performance on the Sleep-EDFx Dataset}
\label{result_fig1}
\end{figure*}

\section{Experiments}
\subsection{Data Collection and Evaluation Metrics}
A band-pass Butterworth filter (1st order) was applied to retain EEG data within the 0.5-35Hz range using the Fpz-Cz channel. The filtered data was then converted into 30-second EEG images, sourced from the Sleep-EDFx dataset \cite{kemp2000analysis,goldberger2000physiobank}. To reduce the cost of generating a large amount of CoT data, 1300 examples from each class of the visualized data were selected for answer generation, yielding a total of 5119 valid CoT-based interpretations (Wake: 1175, N1: 1186, N2: 757, N3: 836, REM: 1165). For model training and evaluation, 75 samples per class were reserved for testing, and the remaining samples were used for training.

Model performance was assessed using a comprehensive set of metrics. 
Class-wise F1-scores were reported to evaluate the model’s ability to distinguish individual sleep stages, particularly those with subtle physiological differences (\textit{e.g.}, N1 and REM). 
Overall performance was further quantified using accuracy (ACC), Cohen’s kappa ($\kappa$), and the macro-averaged F1-score (MF1). 
ACC reflects the overall correctness of the predictions, whereas $\kappa$ accounts for agreement beyond chance and is widely used in clinical sleep scoring. Since our evaluation set contains an equal number of samples per class, MF1 serves as an unbiased estimate of the average per-class performance, ensuring that all sleep stages—especially difficult ones such as N1 and REM—contribute equally to the final assessment.


\subsection{Implementation Details}
A customized ResNet-18 was adopted as the specialized vision module within our visual enhancement Module. 
Two VLM backbones were evaluated: the pre-trained LLaVA-1.5 and the more recent LLaVA-Next.

\textbf{Training the Visual Enhancement Module.}
The customized ResNet-18 was first trained independently for 30 epochs with a learning rate of 5e-4 and a batch size of 8 on an NVIDIA GeForce RTX 4090 GPU to extract high-level semantic representations (\( Z_f \)) from EEG images.

\textbf{Integration with LLaVA-1.5.}
For the LLaVA-1.5-based framework, the trained vision module was integrated into LLaVA-1.5, and the full model was fine-tuned using LoRA-based lightweight adaptation. During joint training, the extracted features \( Z_f \) were projected into the VLM embedding space through a shared projection layer \( W \). LLaVA-1.5-13B was fine-tuned for 2 epochs with a learning rate of 3e-4 on a single NVIDIA A100 GPU, while all other hyperparameters remained at their default settings.

\textbf{Integration with LLaVA-Next.}
To further evaluate the generality of our framework, we also integrated the visual enhancement module into LLaVA-Next. As LLaVA-Next is designed without support for parameter-efficient tuning, the model was fully fine-tuned for 2 epochs on two NVIDIA A100 GPUs. The base language model was set to \textit{Llama-3.1-8B-Instruct}, and all remaining hyperparameters followed the default configuration of LLaVA-Next.
 



\subsection{Baseline Selection}
To comprehensively evaluate the proposed framework, we include several categories of baselines that collectively reflect (i) the native capability of current advanced VLMs on the EEG-based sleep stage classification task, (ii) the task-specific performance achievable by specialized EEG models, and (iii) the contrast between purely visual architectures and VLM-based frameworks.

\textbf{Vision–language models.} 
We evaluate two types of VLM baselines. First, GPT-4 and Qwen2.5-VL-72B are included in their off-the-shelf forms to quantify the native limitations of state-of-the-art VLMs when directly applied to the EEG sleep stage classification task. Second, we incorporate LLaVA-Next-8B as a fine-tuned VLM baseline, trained with full-parameter adaptation to provide a fairer comparison against our proposed approach. Together, these baselines reveal both the inherent challenges faced by general-purpose VLMs and the performance level achievable through conventional VLM adaptation.

\begin{figure*}[t]
    \centering

    \begin{minipage}[c]{0.02\textwidth}
        \centering
        \rotatebox{90}{\small True Stage}
    \end{minipage}
    %
    \begin{minipage}{0.97\textwidth}
        \centering
        \subfloat[Ours-L1.5-R18]{%
            \includegraphics[width=0.25\textwidth]{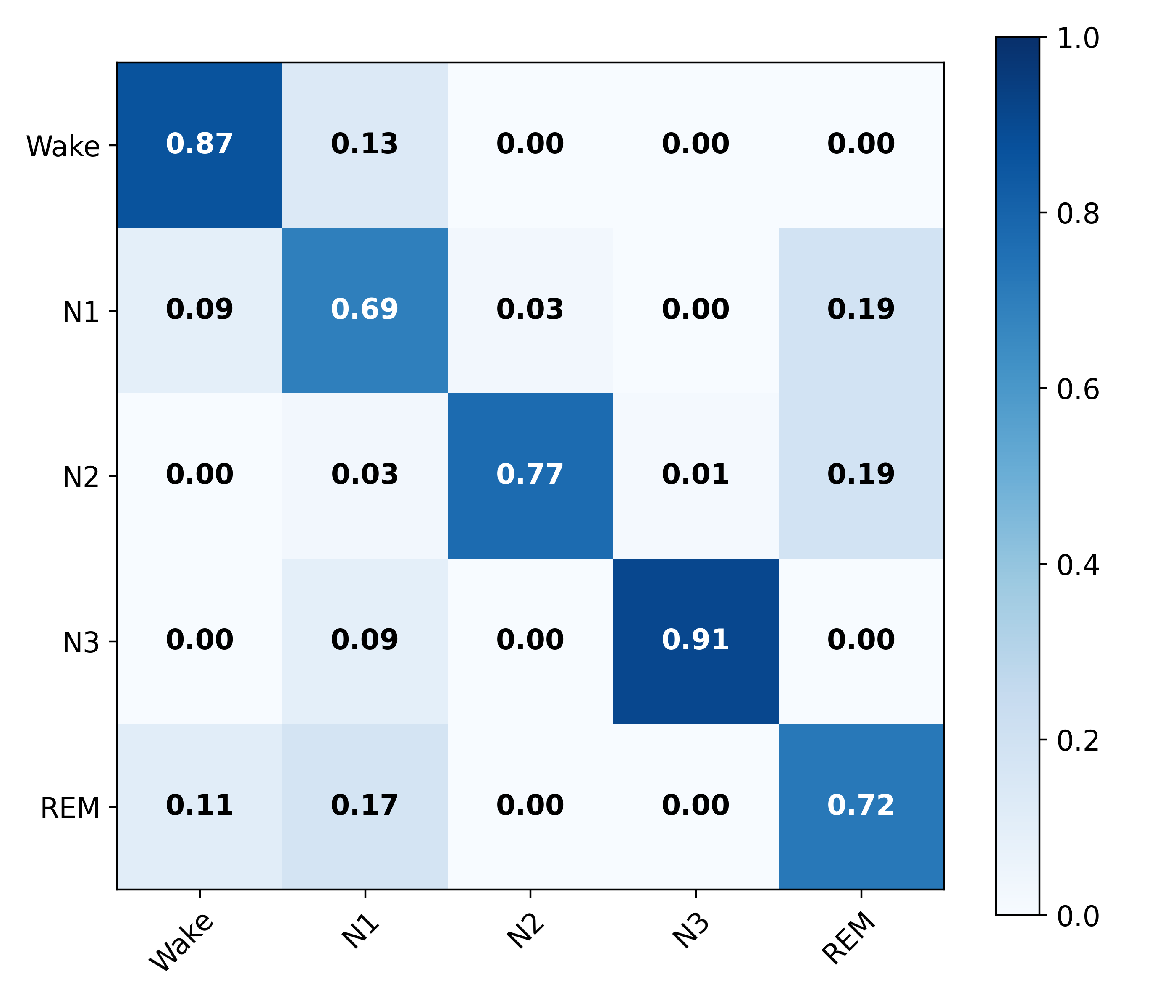}}
        \hfill
        \subfloat[Ours-L1.5-CNx]{%
            \includegraphics[width=0.25\textwidth]{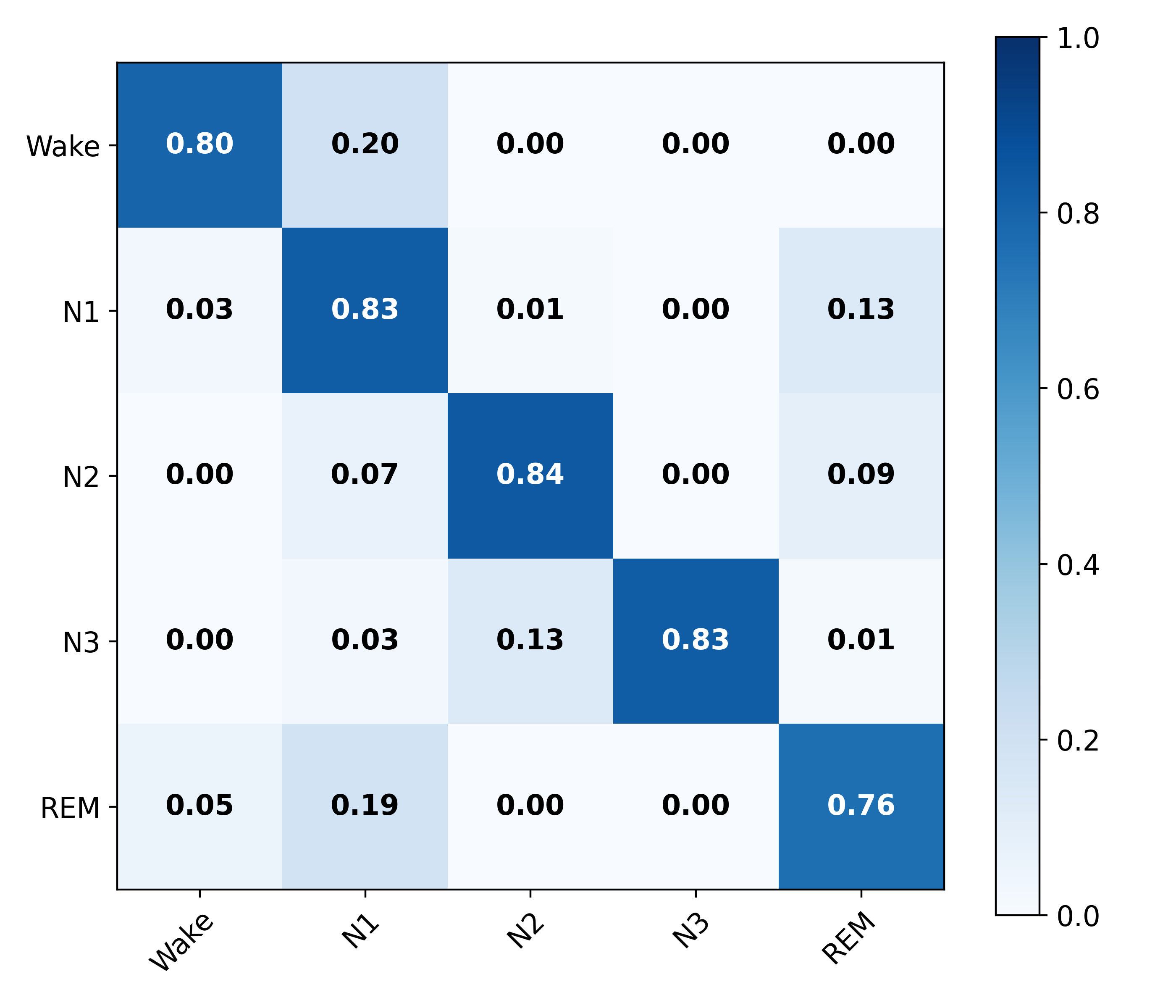}}
        \hfill
        \subfloat[Ours-LNxt-R18]{%
            \includegraphics[width=0.25\textwidth]{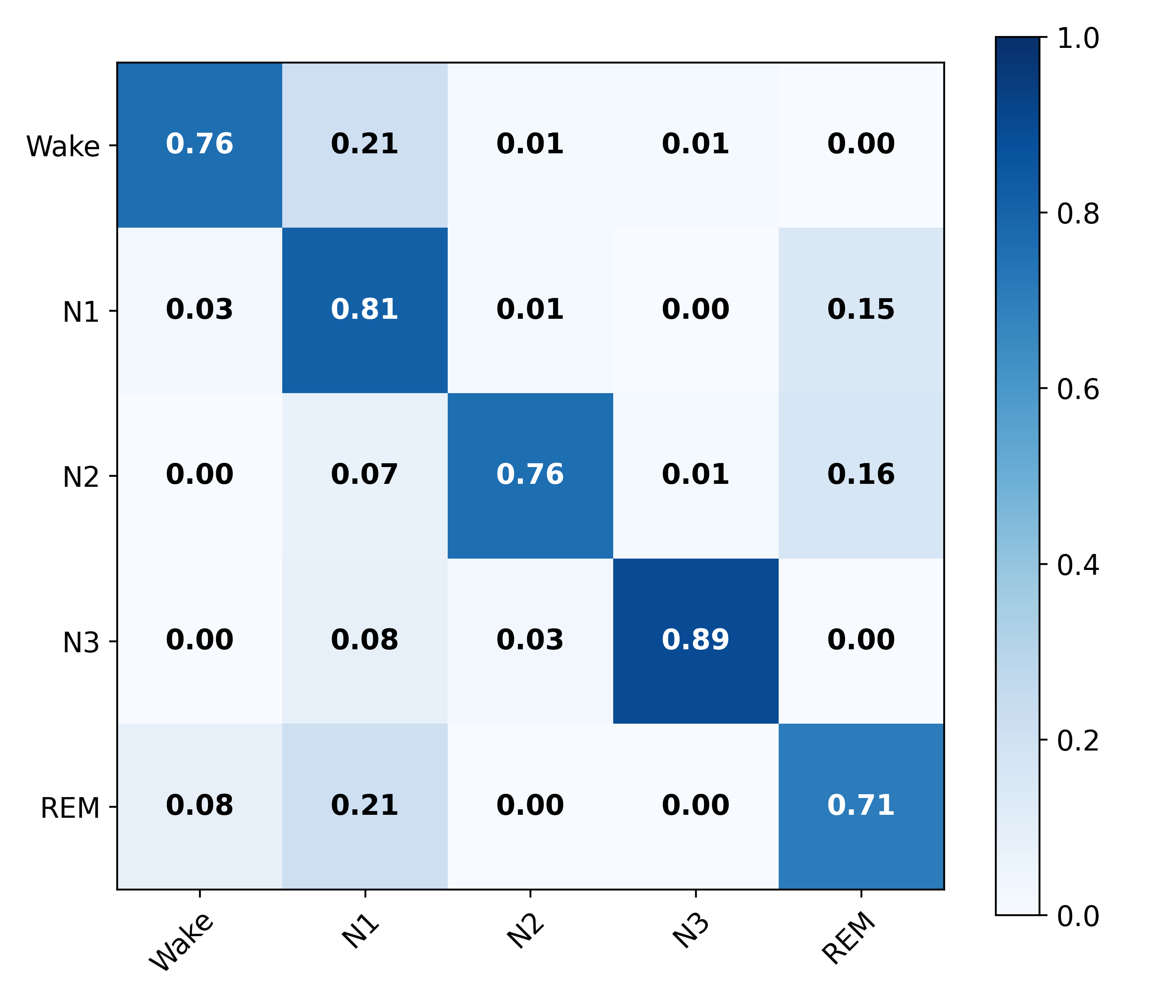}}
        \hfill
        \subfloat[Ours-LNxt-CNx]{%
            \includegraphics[width=0.25\textwidth]{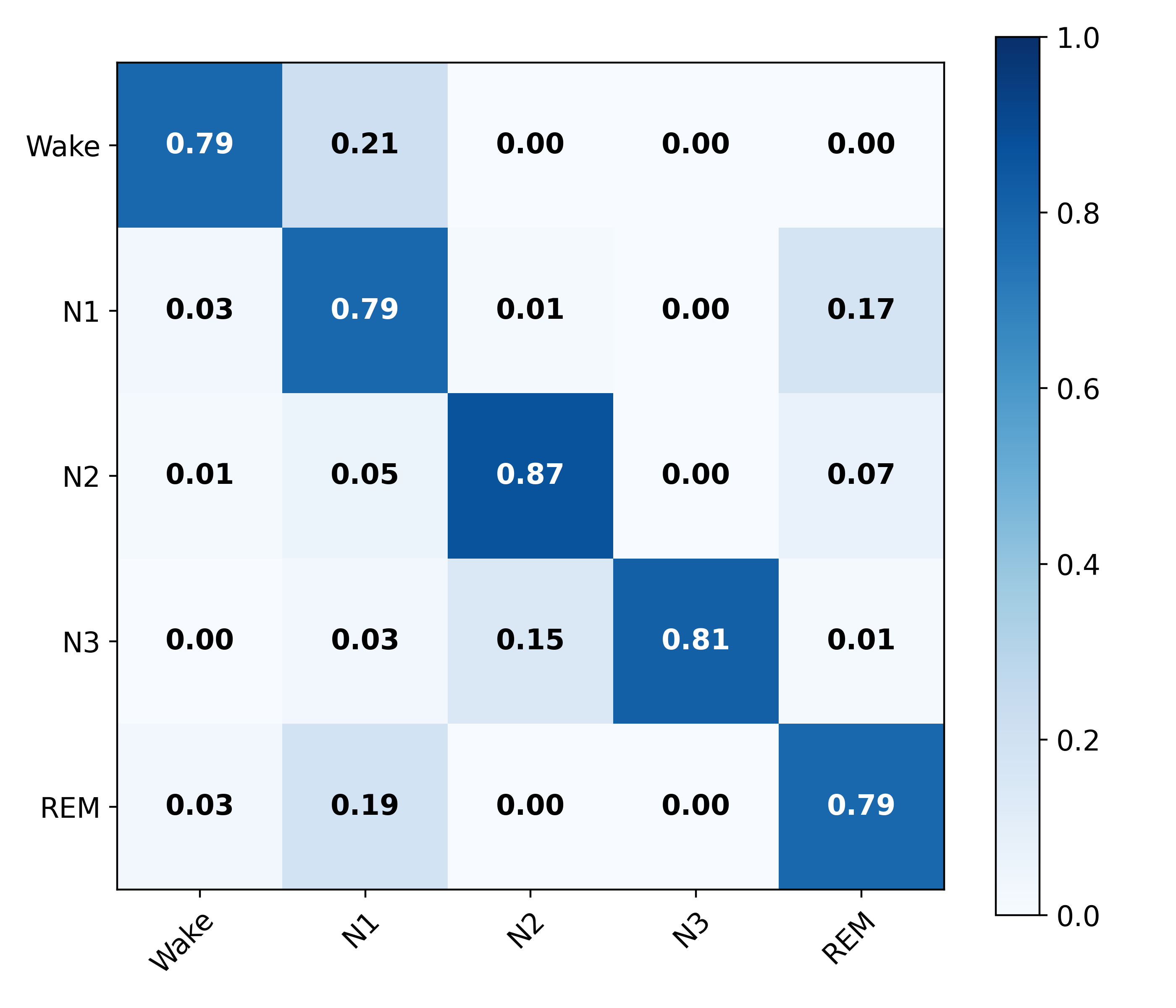}}
    \end{minipage}
    \caption{Confusion matrices of four EEG-VLM model variants on the Sleep-EDFx dataset. Here, \textbf{L1.5} and \textbf{LNxt} denote VLM backbones based on LLaVA-1.5 and LLaVA-Next, while \textbf{R18} and \textbf{CNx} correspond to the ResNet-18 and ConvNeXt-Base backbones used in the specialized vision model.
    }
    \label{ours_confusion}
\end{figure*}

\begin{figure}[tb]
\centering
\includegraphics[width=\columnwidth]{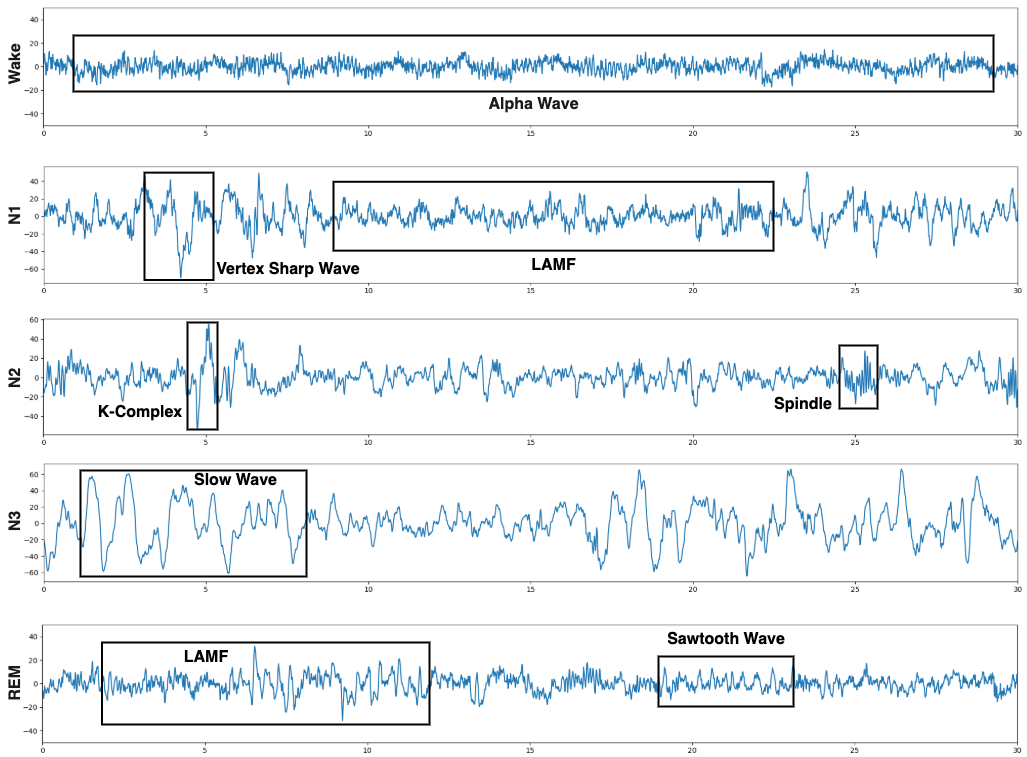}
\caption{Typical EEG characteristics across sleep stages: Wake - Alpha Waves; N1 - Low Amplitude Mixed Frequency (LAMF: Alpha, Beta) and Vertex Sharp Waves; N2 - K-Complexes and Sleep Spindles; N3 - Slow Waves; REM - LAMF (Beta, Theta) and Sawtooth Waves.}
\label{wave-demo}
\end{figure}

\textbf{Specialized EEG model.} 
FFTCN~\cite{bao2024feature} is included as a strong and representative deep learning model designed specifically for single-channel EEG sleep stage classification. This baseline provides a task-specific performance reference, allowing us to contextualize the effectiveness and potential advantages of VLM-based approaches relative to specialized architectures.

\textbf{Conventional vision backbones.} 
We additionally include ResNet-18 and ConvNeXt-Base as representative standalone vision backbones. These models are not intended as direct competitors; rather, they serve as a reference for contrasting traditional vision architectures with VLM-based models.

\subsection{Main Results}\label{sub:results}
The quantitative results on the Sleep-EDFx dataset are presented in Table~\ref{tab:Main} and Fig.~\ref{result_fig1}. We observe that off-the-shelf VLMs—despite their strong general multimodal capabilities—perform poorly when directly applied to EEG sleep staging. GPT-4 and Qwen2.5-VL-72B achieve accuracies of only 0.205 and 0.243, with Kappa scores close to zero, indicating almost no agreement with the clinical ground truth. After full-parameter adaptation, LLaVA-Next-8B shows a notable improvement (Accuracy = 0.533, MF1 = 0.483, Kappa = 0.417), but its overall performance remains far below the level required for reliable clinical use. These results highlight the intrinsic difficulty of modeling fine-grained EEG patterns and the limitations of current VLM architectures in capturing frequency–amplitude structures essential for sleep staging.

In contrast, our proposed EEG-VLM framework substantially elevates VLM performance. Among all variants, Ours-L1.5-CNx achieves the best overall results (Accuracy = 0.811, MF1 = 0.816, Kappa = 0.763), while the remaining variants also show consistent and robust gains. These improvements arise from the framework’s hierarchical visual enhancement and structured reasoning design, which together strengthen the model’s fine-grained visual perception and domain-specific inference capability, enabling accurate and clinically interpretable EEG-based sleep staging.

Although FFTCN achieves slightly higher overall accuracy, its design objectives differ fundamentally from ours. Our goal is not to surpass all specialized EEG models on every metric, but to demonstrate the feasibility, unique advantages, and added interpretability of a VLM-based paradigm for physiological waveform analysis.

\begin{table*}[t]
\centering
\renewcommand{\arraystretch}{1.15}
\setlength{\tabcolsep}{7pt}

\caption{Ablation Study: Exploring Embedding and Reasoning Strategies. 
\textbf{Bold} indicates the best result and \underline{underline} indicates the second-best result.}
\label{tab:Ablation}

\rowcolors{2}{}{gray!8}

\begin{tabular}{c ccc ccccc}
\toprule
\textbf{Configurations} &
\multicolumn{3}{c}{\textbf{Overall Results}} &
\multicolumn{5}{c}{\textbf{F1-score for Each Class}} \\
\cmidrule(lr){2-4} \cmidrule(lr){5-9}

 & \textbf{Accuracy} & \textbf{MF1} & \textbf{Kappa} &
   \textbf{Wake} & \textbf{N1} & \textbf{N2} & \textbf{N3} & \textbf{REM} \\
\midrule

W/O Feature Embedding 
    & 0.271 & 0.181 & 0.085 & 0.377 & 0.280 & 0.000 & 0.000 & 0.247 \\

Raw \( H_f \) Embedding 
    & 0.264 & 0.153 & 0.080 & 0.387 & 0.026 & 0.000 & 0.000 & 0.351 \\

Patch-Aligned \( H_f \) to \( H_v \) 
    & \underline{0.784} & \underline{0.789} & \underline{0.730} 
    & \underline{0.841} & \underline{0.659} & \underline{0.857} 
    & \textbf{0.944} & \underline{0.645} \\

W/O CoT Reasoning 
    & 0.728 & 0.735 & 0.660 & 0.800 & 0.584 & 0.836 & 0.922 & 0.533 \\

GPT-4 Analysis 
    & 0.757 & 0.761 & 0.697 & 0.824 & 0.624 & 0.840 & 0.730 & 0.582 \\

Label-Guided Pre-Analysis 
    & 0.621 & 0.598 & 0.527 & 0.749 & 0.533 & 0.348 & \underline{0.937} & 0.426 \\

\rowcolor{white}
\textbf{Ours-L1.5-R18} 
    & \textbf{0.792} & \textbf{0.797} & \textbf{0.740} 
    & \textbf{0.839} & \textbf{0.654} & \textbf{0.859} 
    & \textbf{0.944} & \textbf{0.688} \\

\bottomrule
\end{tabular}
\end{table*}

\begin{figure*}[tb]
\centering
\includegraphics[width=0.90\textwidth]{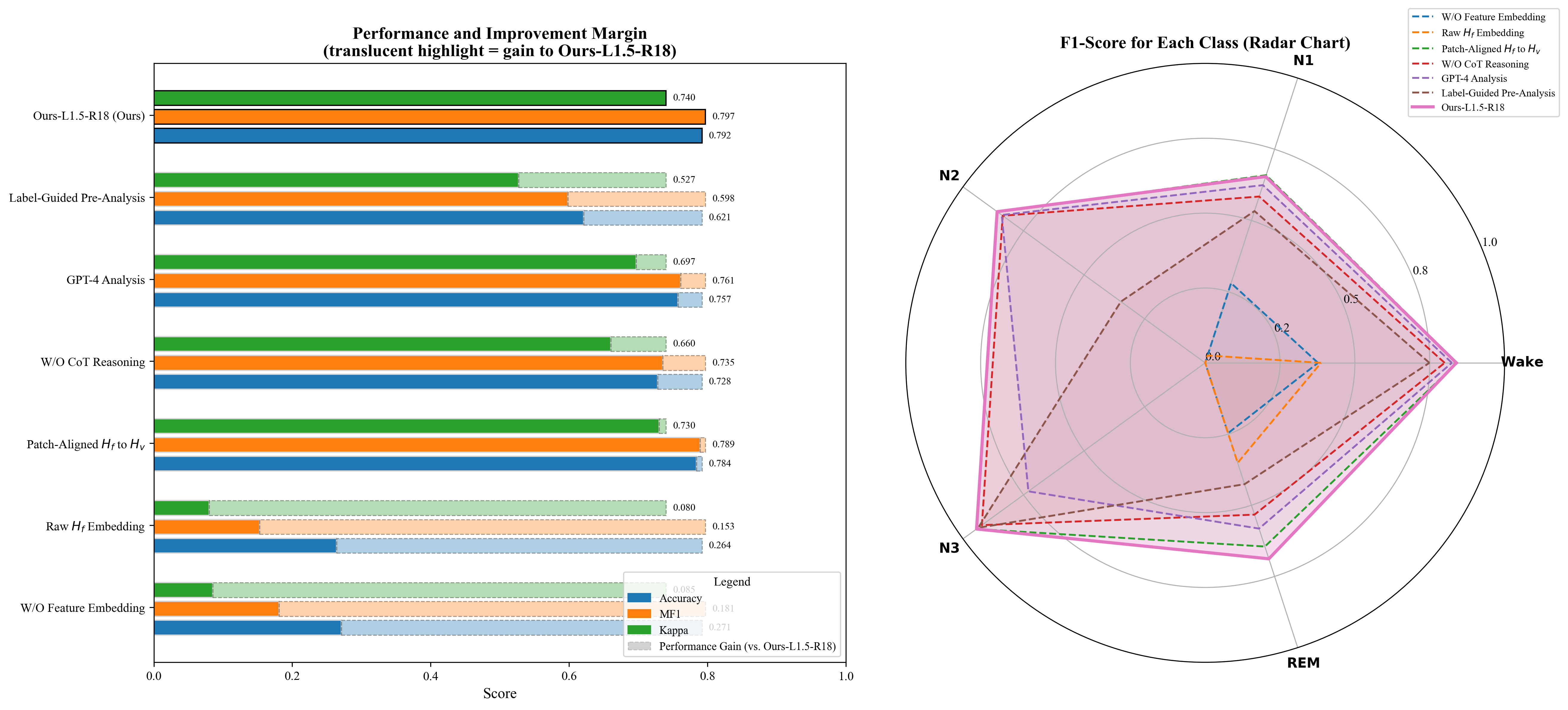}
\caption{Ablation Study of Feature Embedding and Reasoning Strategies} \label{result_fig2}
\end{figure*}

Beyond overall metrics, a more detailed examination reveals that our method offers more stable performance on physiologically ambiguous stages, particularly Wake, N1, and REM, where waveform morphology and frequency content overlap substantially. As shown in Fig.~\ref{ours_confusion}, all four EEG-VLM variants achieve notably higher accuracy on the N1 stage and exhibit fewer misclassifications. This improvement stems from our model’s cross-modal semantic enhancement and stage-wise CoT reasoning, which emulate the step-by-step diagnostic process used by human experts.

These stages are inherently difficult to discriminate due to shared waveform structures. As illustrated in Fig.~\ref{wave-demo}, both Wake and N1 contain alpha activity, N1 and REM share low-amplitude mixed-frequency (LAMF) components with beta involvement, and both Wake and REM exhibit prominent low-frequency content. Consequently, even strong EEG-specific models such as FFTCN may misclassify N1 due to these overlapping morphological characteristics.

By comparison, our EEG-VLM framework mitigates such confusion by incorporating hierarchical visual representation and interpretable reasoning, ultimately improving robustness on ambiguous stages. These results highlight the promise of VLM-based approaches for physiological waveform interpretation and their potential as a complementary paradigm to conventional deep learning models.

As standalone vision backbones, ResNet-18 and ConvNeXt-Base achieve competitive accuracies of 0.752 and 0.813, demonstrating the effectiveness of modern visual architectures in extracting discriminative features from EEG images. However, these purely visual models lack the ability to model stage-specific semantics and clinical context. In contrast, our visual enhancement module introduces high-level semantic representations from intermediate layers, enabling the VLM to achieve stronger fine-grained representation and reasoning capability.

Although the LLaVA-Next–based variants exhibit strong performance, they remain slightly inferior to the LLaVA-1.5–based counterparts. We conjecture that this gap primarily arises from two factors. First, the LLaVA-1.5 configuration in our framework uses a larger 13B language backbone, whereas LLaVA-Next is restricted to an 8B model, which limits its capacity to capture fine-grained EEG patterns under the same data regime. Second, LLaVA-1.5 is adapted via lightweight LoRA tuning, preserving much of its pretrained multimodal alignment and reasoning capability, while LLaVA-Next undergoes full-parameter fine-tuning on a relatively small EEG dataset and is therefore more susceptible to overfitting and partial forgetting of its generic visual–language knowledge. These factors jointly explain the modest performance gap observed between the two model families.

\begin{table*}[t]
\centering
\renewcommand{\arraystretch}{1.1}
\setlength{\tabcolsep}{7pt}

\caption{Performance Evaluation on Data from a Collaborative Hospital. 
\textbf{Bold} indicates the best result and \underline{underline} indicates the second-best result.}
\label{tab:External}

\rowcolors{3}{}{gray!8}

\begin{tabular}{c ccc ccccc}
\toprule

\textbf{Method} &
\multicolumn{3}{c}{\textbf{Overall Results}} &
\multicolumn{5}{c}{\textbf{F1-score for Each Class}} \\
\cmidrule(lr){2-4} \cmidrule(lr){5-9}

 & \textbf{Accuracy} & \textbf{MF1} & \textbf{Kappa} &
   \textbf{Wake} & \textbf{N1} & \textbf{N2} & \textbf{N3} & \textbf{REM} \\
\midrule

ResNet-18 
    & 0.674 & 0.675 & 0.592 
    & 0.646 & 0.591 & 0.828 & \underline{0.819} & 0.489 \\

Patch-Aligned-R18
    & \underline{0.710} & \underline{0.717} & \underline{0.638}
    & 0.660 & \underline{0.601} & \underline{0.832} & 0.868 & \underline{0.623} \\

ConvNeXt-Base
    & 0.702 & 0.710 & 0.627
    & \underline{0.712} & 0.610 & 0.745 & \underline{0.819} & 0.662 \\

\textbf{Ours-L1.5-R18}
    & \textbf{0.751} & \textbf{0.756} & \textbf{0.689}
    & 0.697 & \textbf{0.638} & \textbf{0.873} & \textbf{0.886} & \underline{0.688} \\

\textbf{Ours-L1.5-CNx}
    & 0.719 & 0.722 & 0.649
    & \textbf{0.752} & 0.632 & 0.641 & 0.857 & \textbf{0.727} \\

\bottomrule
\end{tabular}
\end{table*}

\begin{figure*}[tb]
\centering
\includegraphics[width=0.90\textwidth]{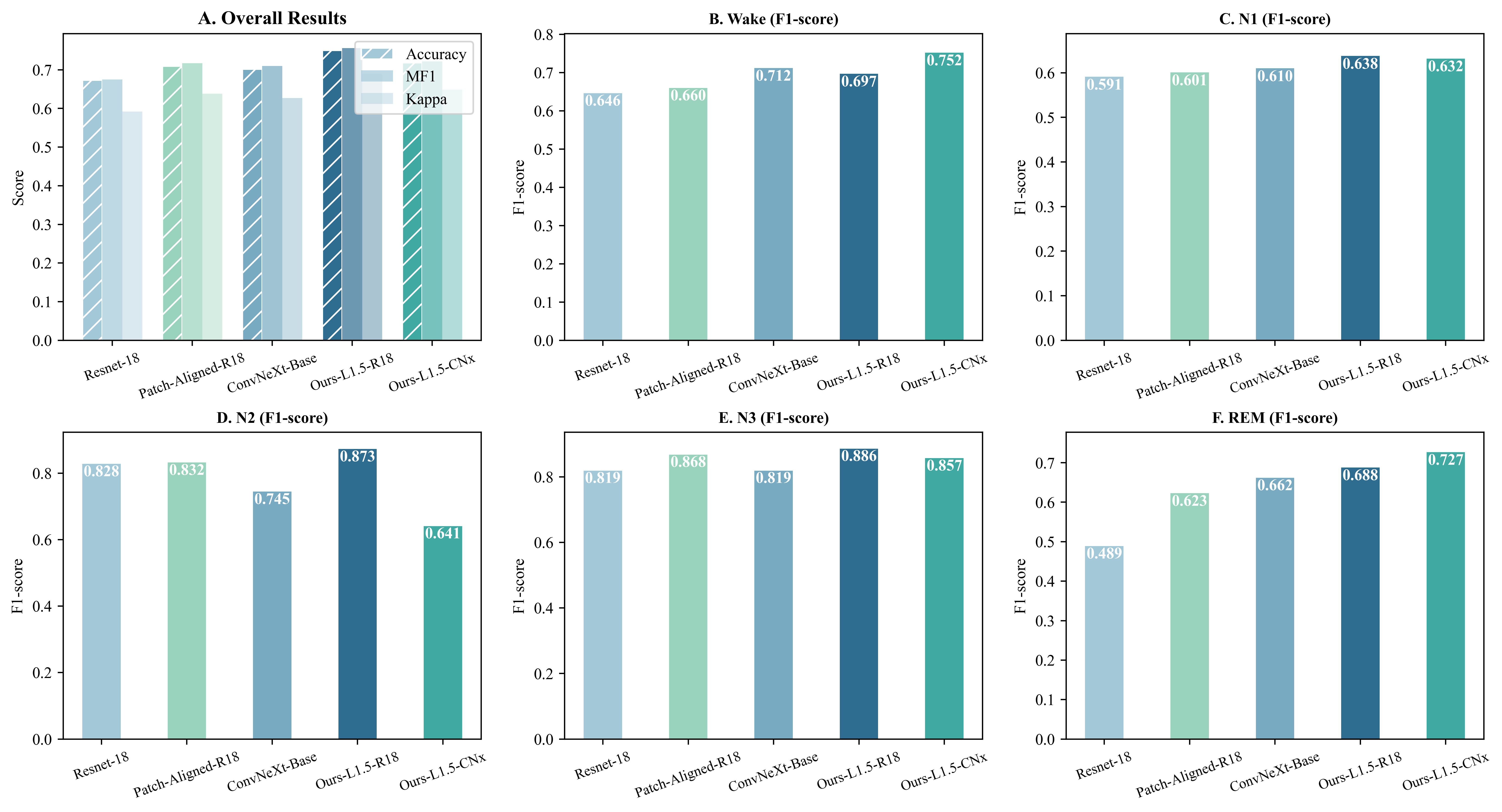}
\caption{Overall and Per-Stage Classification Performance on the Collaborative Hospital's EEG Dataset
} 
\label{result_fig3}
\end{figure*}

\subsection{Ablation Study}
We conduct a comprehensive ablation study on the \textbf{Ours-L1.5-R18} variant, and the results in Table~\ref{tab:Ablation} and Fig.~\ref{result_fig2} reveal the contribution of each component. First, removing the \textit{Feature Embedding} leads to a substantial performance collapse, consistent with the weaker results observed for LLaVA-Next-8B in Section~\ref{sub:results}. This confirms that incorporating hierarchical visual representations is essential for equipping VLMs with sufficient visual understanding of EEG images. Similarly, using raw \(H_f\) features alone results in poor performance, whereas applying \textit{Patch-Aligned \(H_f\) to \(H_v\)} produces a significant improvement. This validates the necessity of multi-level feature alignment, which enables \(H_v\) to capture both the fine-grained morphological details and the global semantic information in \(H_f\).

Furthermore, removing \textit{CoT Reasoning} causes a clear performance drop, demonstrating that structured inference plays a crucial role in both interpretability and decision quality. Interestingly, the \textit{GPT-4 Analysis} configuration confirms the independent contribution of GPT-generated CoT reasoning, while the 
\textit{Label-Guided Pre-Analysis} variant underperforms. This suggests that injecting label information before reasoning may disrupt the structured inference process rather than enhance it.

Overall, our full model—\textbf{Ours-L1.5-R18}, which integrates hierarchical representation learning with an optimized CoT prompting strategy—achieves the best performance, confirming the effectiveness and synergy of all proposed components within the EEG-VLM framework.

\subsection{External Validation}
To assess the cross-dataset generalization capability of our model, we conducted evaluation on an external EEG dataset collected from the collaborative university hospital. Since this dataset does not include the Fpz-Cz channel, we used the C4-M1 channel for testing and applied the same preprocessing procedure as used for the Sleep-EDFx dataset, sampling 250 examples per class.

Previous experiments have shown that generic VLM baselines perform poorly on EEG sleep stage classification. Moreover, we have demonstrated that, compared with the specialized deep learning model FFTCN, our approach provides more stable performance on physiologically ambiguous stages such as Wake, N1, and REM, further confirming the feasibility and unique advantages of a VLM-based paradigm for physiological waveform analysis. Therefore, the external validation focuses primarily on CNN backbones, ablation variants, and different configurations of the proposed model.

As shown in Table~\ref{tab:External} and Fig.~\ref{result_fig3}, even under different channel configurations and higher noise levels, both Ours-L1.5-R18 and Ours-L1.5-CNx exhibit strong cross-dataset robustness, with particularly stable performance on the REM stage. Unlike the main results on Sleep-EDFx, Ours-L1.5-R18 slightly outperforms Ours-L1.5-CNx on this external dataset. We attribute this to the simpler architecture of ResNet-18, which tends to generalize better across domains, whereas ConvNeXt is more sensitive to shifts in data distribution and can be more affected by morphological and noise variations across channels and recording environments.

Overall, our method clearly outperforms both ResNet-18 and ConvNeXt-Base, indicating that hierarchical visual representation and structured reasoning provide advantages beyond what purely visual models can achieve. This combination enhances robustness to channel shifts and noise variations, leading to better cross-dataset stability and generalization.

\section{Discussion}

This study demonstrates the feasibility and advantages of integrating vision–language models with hierarchical visual representation learning and CoT-based reasoning for EEG-based sleep stage classification. Although VLMs have achieved remarkable success in natural image and general multimodal tasks, our results show that their native capability does not directly transfer to physiological waveform interpretation, largely due to insufficient fine-grained visual perception and limited domain-specific reasoning.

By incorporating a specialized visual enhancement module, a multi-level feature alignment mechanism, and a CoT-guided reasoning strategy, the proposed EEG-VLM framework substantially improves both accuracy and interpretability. The visual enhancement module provides high-level semantic cues tailored to EEG morphology, compensating for the limited visual expressiveness of CLIP-based encoders. Multi-level alignment further enables the combination of global and local visual information, allowing more precise discrimination of subtle waveform patterns. In addition, stage-wise CoT prompting introduces clinically meaningful reasoning steps, reducing confusion among physiologically similar stages and offering interpretable insights into the decision process.

A notable finding is the superior robustness of our framework on ambiguous sleep stages such as Wake, N1, and REM—categories that remain challenging even for specialized EEG models. This highlights the potential of combining enhanced visual semantics with structured reasoning to capture subtle physiological patterns. The cross-dataset evaluation further confirms the generalization capability of our method. Despite differences in channels, recording environments, and noise characteristics, the proposed models maintain stable performance. We also observe that ResNet-based variants generalize better than their ConvNeXt-based counterparts, suggesting that simpler visual architectures may exhibit stronger cross-domain robustness when integrated with hierarchical VLM reasoning. These observations provide practical guidance for designing VLM extensions aimed at real-world clinical settings, where data heterogeneity is unavoidable.

Despite these promising results, several limitations remain. First, our framework adopts a modified ResNet-18 as the specialized vision model backbone. Future work could explore transformer-based or hybrid convolution–attention architectures to improve global–local feature modeling in EEG images. Second, although the proposed patch-aligned fusion strategy is simple and effective, it still incurs additional computational cost. Adaptive fusion mechanisms such as low-rank decomposition may further improve efficiency. Third, our study focuses on single-channel EEG, while additional physiological signals (\textit{e.g.}, EOG, EMG) may provide richer spatial and contextual information. Incorporating such modalities into a unified vision–language framework represents an important direction for future research.

In summary, this work provides strong evidence that VLM-based frameworks, when equipped with appropriate visual enhancement and structured reasoning mechanisms, offer a powerful and interpretable paradigm for physiological waveform analysis. We believe that EEG-VLM lays a foundation for extending multimodal large models to broader electrophysiological and clinical decision-support applications.

\section{Conclusion}
In this study, we introduced a hierarchical vision–language framework that improves EEG image-based sleep stage classification through multi-level feature alignment and visually enhanced language-guided reasoning. The proposed method leverages a visual enhancement module to extract high-level semantic representations from intermediate visual features and integrates them with low-level CLIP embeddings via a multi-level alignment mechanism. In addition, CoT-based reasoning enables interpretable, step-wise inference that simulates expert decision-making, enhancing both visual understanding and domain-specific reasoning within VLMs. Experimental results demonstrate that the proposed framework achieves excellent performance and generalization across multiple datasets, with notable advantages in transparency and interpretability. In particular, the method exhibits superior robustness in challenging stages such as N1 and REM, where traditional models often struggle. We believe this work provides a solid foundation for expanding VLM-based approaches to broader physiological signal analysis and future clinical decision-support applications.

\section*{References}

\def\refname{\vadjust{\vspace*{-2.5em}}} 

\end{document}